# Cross-Lingual Response Consistency in Large Language Models: An ILR-Informed Evaluation of Claude Across Six Languages

**Camelia Baluta**

*Independent Researcher | Former Faculty, Defense Language Institute Foreign Language Center*
*ILR/OPI Assessment Specialist*
April 2026

## Abstract

Large language models (LLMs) trained on multilingual corpora do not treat all languages equally. Training data distributions, tokenization artifacts, and reinforcement learning from human feedback sourced predominantly from English-speaking annotators may produce outputs that vary in quality, length, pragmatic calibration, and cultural framing across languages — even when the semantic input is held constant. This paper introduces a systematic evaluation framework grounded in the Interagency Language Roundtable (ILR) Skill Level Descriptions and applies it to Claude (Sonnet 4.6) across six languages: English, French, Romanian, Spanish, Italian, and German. We administer a battery of 12 semantically equivalent prompt clusters spanning ILR complexity levels 1 through 3+, collect 216 responses (12 prompts × 6 languages × 3 runs), and analyze outputs through a two-layer methodology combining automated quantitative metrics with expert ILR qualitative assessment. Quantitative analysis reveals that French responses are approximately 30% longer than German responses on identical prompts, and that creative and affective clusters show the highest cross-lingual surface divergence. Qualitative analysis — conducted by a six-language professional with 12 years of ILR/OPI assessment experience — identifies five cross-lingual variation patterns: systematic differences in pragmatic disambiguation strategies, aesthetic and literary tradition divergence in creative output, language-internal technical terminology norms, cultural calibration gaps evidenced by the absence of culture-specific content in favor of culturally neutralized templates, and language-specific institutional referral behavior in emotional support responses. We argue that ILR-informed expert judgment applied to LLM outputs constitutes a novel and underreported evaluation methodology that complements purely computational benchmarks, and that cross-lingual output variation in Claude is interpretable, domain-dependent, and consequential for equitable multilingual AI deployment.

# 1. Introduction

## 1.1 The Problem of Cross-Lingual Consistency

When a user asks Claude a question in Romanian, do they receive a response of equivalent quality, cultural appropriateness, and pragmatic calibration to a user asking the same question in English? This question has significant implications for the equitable deployment of AI systems across language communities — yet it remains underexplored in the literature.

Existing cross-lingual LLM benchmarks typically measure factual accuracy, reasoning correctness, or task completion on structured tasks (Ahuja et al., 2023; Bang et al., 2023; Lai et al., 2023). These approaches capture important dimensions of model performance but are less equipped to measure the pragmatic, stylistic, and culturally embedded dimensions of language use that determine whether a response feels appropriate, complete, and well-calibrated to its linguistic context. A response can be factually correct and pragmatically inadequate simultaneously — a distinction that structured benchmarks are not designed to capture.

This gap is particularly significant for languages that are well-represented in training data but whose cultural and pragmatic conventions may be underrepresented relative to English. French, German, Spanish, Italian, and Romanian are all high-resource European languages with substantial training data presence — yet their speakers may still receive outputs that are calibrated to English-language discourse norms rather than the conventions of their own language communities.

## 1.2 The ILR Framework as an Evaluation Tool

The Interagency Language Roundtable (ILR) Skill Level Descriptions (2012) provide a principled, field-tested framework for describing linguistic and pragmatic competence across five skill areas at levels 0 through 5. Developed for human language assessment and widely used in U.S. government language training and certification, the ILR scale has not previously been applied systematically to LLM output evaluation.

This project proposes and operationalizes such an application. We introduce two constructs adapted from the ILR framework for LLM output analysis:

**Prompt Complexity Level (PCL):** The ILR level a human would need to fully comprehend and respond appropriately to a given prompt, annotated by an expert assessor for each prompt in the battery.

**Response Adequacy Level (RAL):** The ILR level a human would need to have produced the observed response — a measure of the linguistic and pragmatic sophistication of the model's output assessed against the demands of the prompt that generated it.

The central research question is: *Does RAL vary significantly across languages when PCL is held constant?*

## 1.3 Why This Project and Why Now

Several convergent factors motivate this study at this moment.

First, frontier LLMs are being deployed at scale across multilingual user populations with limited systematic evaluation of cross-lingual output consistency. The consequences of inconsistency are not merely academic — they include differential quality of AI-assisted services across language communities, potential cultural misrepresentation, and the reproduction of English-centric discourse norms in non-English contexts.

Second, the ILR framework provides a uniquely appropriate vocabulary for this evaluation. Unlike automated metrics that measure surface properties, ILR descriptors capture the functional and pragmatic dimensions of language use — precisely the dimensions most likely to vary meaningfully across languages in LLM outputs.

Third, this project is conducted by a researcher with direct expertise in ILR/OPI assessment across all six target languages, enabling a quality of qualitative judgment that purely computational approaches cannot replicate. The researcher's ILR 4+ Romanian competence is particularly relevant for identifying the absence of culture-specific content — a finding that requires insider knowledge to make credibly.

### 1.4 Scope and Contributions

This paper makes four primary contributions:

1. **A novel evaluation framework** adapting ILR Skill Level Descriptions to LLM output analysis through the constructs of Prompt Complexity Level (PCL) and Response Adequacy Level (RAL).
2. **A citable prompt battery** of 12 semantically equivalent prompt clusters across six languages, annotated with ILR complexity levels, calibration rationale, linguistic feature targets, and researcher hypotheses, made publicly available at github.com/camelbal-ship-it/crosslingual-claude-eval.
3. **Five cross-lingual variation findings** in Claude (Sonnet 4.6) that are systematic, interpretable, and theoretically coherent with known cross-cultural pragmatic differences.
4. **A methodological argument** for the value of expert ILR judgment in LLM evaluation — a contribution to the growing conversation about what human expertise can provide that automated metrics cannot.

### 1.5 Paper Organization

Section 2 reviews related work in cross-lingual LLM evaluation and identifies the gap this project addresses. Section 3 describes the methodology, including prompt battery design, data collection protocol, and the two-layer analysis framework. Section 4 presents quantitative and qualitative results. Section 5 discusses implications for multilingual AI deployment and evaluation methodology. Section 6 acknowledges limitations and outlines directions for future work.

## 2. Related Work

## 2.1 Cross-Lingual LLM Benchmarks

Several recent studies have evaluated LLM performance across multiple languages. Bang et al. (2023) evaluated ChatGPT on 37 tasks across 26 languages, finding significant performance degradation for low-resource languages. Ahuja et al. (2023) introduced MEGA, a multilingual evaluation benchmark covering 16 NLP datasets across 70 languages. Lai et al. (2023) assessed cross-lingual reasoning and generation tasks across multiple model families. These studies established that cross-lingual performance gaps exist and are substantial for many languages.

However, these benchmarks share a common design assumption: that cross-lingual quality can be measured through task accuracy on structured inputs with ground-truth answers. This assumption is appropriate for factual recall, mathematical reasoning, and classification tasks, but systematically underspecifies the evaluation of pragmatic, stylistic, and culturally embedded language use — precisely the dimensions most relevant to user experience in open-ended conversational contexts.

## 2.2 Pragmatic and Cultural Evaluation

A smaller body of work addresses cultural and pragmatic dimensions of LLM outputs. Hershcovich et al. (2022) introduced a framework for cross-cultural NLP, arguing that current systems insufficiently account for cultural variation and calling for more culturally-aware evaluation and modeling approaches. Cao et al. (2023) probed cultural knowledge in LLMs across multiple languages, finding that models demonstrate uneven cultural knowledge with English-centric biases.

None of these studies apply a proficiency-based framework to evaluate the linguistic and pragmatic adequacy of LLM outputs against the functional demands of the prompts that generated them. The ILR framework fills this gap.

## 2.3 The ILR Framework in Language Assessment

The ILR Skill Level Descriptions (Interagency Language Roundtable, 2012) are the standard proficiency framework for U.S. government language assessment, used in hiring, training evaluation, and certification across military, intelligence, and diplomatic contexts. The Oral Proficiency Interview (OPI) and its written equivalent (WPT) are the primary instruments for ILR assessment, conducted by certified assessors.

The ILR scale has been applied to machine translation evaluation through the ARPA MT program, which used ILR proficiency levels to calibrate text difficulty and output quality assessment (White, O'Connell, & O'Mara, 1994), and to calibrate language learning materials. Related work exists on using CEFR proficiency levels to control LLM output complexity for language learning applications (Malik et al., 2024), and on mapping model outputs to proficiency frameworks such as CEFR and ACTFL. However, ILR-level annotation is not a standard or widely adopted evaluation framework for LLM-generated outputs — it is rarely used as a primary annotation schema in LLM evaluation literature, and no published study has applied it systematically to cross-lingual output consistency analysis. This paper proposes and operationalizes that application, arguing that the ILR framework's functional and pragmatic descriptors are particularly well-suited to capturing the dimensions of language use most likely to vary meaningfully across languages in open-ended conversational AI outputs.

## 3. Methodology

### 3.1 Background and Motivation

Large language models (LLMs) trained on multilingual corpora do not treat all languages equally. Training data distributions, tokenization artifacts, and RLHF feedback sourced predominantly from English-speaking annotators may produce outputs that vary in quality, length, hedging behavior, and cultural framing across languages — even when the semantic input is held constant.

This project introduces a systematic evaluation framework grounded in the **Interagency Language Roundtable (ILR) Skill Level Descriptions** — a proficiency framework developed for human language assessment — and adapts it for the analysis of LLM-generated outputs. The ILR framework provides a principled vocabulary for describing linguistic complexity, register, and pragmatic adequacy that is well-suited to the cross-lingual comparison task.

### 3.2 The ILR Framework: Standard Application

The ILR scale (Levels 0–5, with "+" subdivisions) describes human language proficiency across five skill areas: Speaking, Listening, Reading, Writing, and Translation. Level 3 ("Professional Working Proficiency") is the operational threshold for professional communication; Level 5 ("Functionally Native") represents near-native performance.

**ILR Skill Level Descriptors (abbreviated):**

| Level | Label | Core Descriptor |
|---|---|---|
| 0 | No Proficiency | No functional ability |
| 1 | Elementary | Basic survival-level communication |
| 2 | Limited Working | Routine social and simple professional communication |
| 2+ | Limited Working+ | Ability to cover most common topics, beginning of abstract discussion |
| 3 | Professional Working | Full range of professional topics, some nuance |
| 3+ | Professional Working+ | Sophisticated argumentation, precision |
| 4 | Full Professional | Near-complete command; subtle register control |
| 4+ | Full Professional+ | Highly specialized or technical discourse |
| 5 | Distinguished | Equivalent to a highly educated native speaker, erudite across domains |

## 3.3 Novel Application: LLM Output Analysis

### 3.3.1 Reframing the Framework

This project applies ILR descriptors to **model outputs**, not human speakers. The central research question is:

> *Given semantically equivalent prompts at known ILR complexity levels, do Claude's outputs vary in linguistic level and pragmatic calibration across languages — and if so, in what patterns?*

To answer this, we introduce two constructs:

**Prompt Complexity Level (PCL):** The ILR level a human would need to produce the prompt text and fully comprehend its pragmatic demands. Annotated per prompt in prompt_battery.json by the researcher based on ILR/OPI assessment expertise.

**Response Adequacy Level (RAL):** The ILR level a human would need to have produced the observed response — a measure of the linguistic and pragmatic sophistication of the model's output in a given language, assessed against the demands of the prompt that generated it.

The research question becomes operationally: **Does RAL vary significantly across languages when PCL is held constant?**

### 3.3.2 Two-Layer Analysis Design

The project uses a two-layer analysis structure that reflects the nature of the research question and the researcher's expertise.

**Layer 1 — Quantitative Consistency Analysis (automated)**

The first layer establishes *whether* variation exists and provides measurable, reproducible evidence of its presence or absence. Metrics include:

| Metric | What It Measures | Tool |
|---|---|---|
| Response length (tokens, words) | Surface-level output volume differences | collect_responses.py / length_analysis.py |
| Sentence count and mean length | Structural elaboration | preprocess.py |
| Hedging marker density | Epistemic caution expression | Language-specific lexicon lookup |
| Surface similarity to EN baseline | Degree of lexical overlap across languages | TF-IDF character n-gram cosine similarity (semantic_similarity.py) |
| List vs. prose structure | Organizational formatting choices | Pattern detection |

These metrics answer: *Is there variation, and where is it largest?*

**Layer 2 — Qualitative Pragmatic Assessment (expert judgment)**

The second layer answers *what the variation means* — whether it represents a failure of pragmatic calibration, a culturally appropriate adaptation, or a genuine adequacy gap relative to the ILR level the prompt demands.

This layer draws directly on the researcher's ILR/OPI assessor background and six-language proficiency. For each prompt cluster, responses across all six languages are reviewed for:

- **Register appropriateness:** Does the response match the formality level the prompt context implies?
- **Pragmatic completeness:** Does the response fulfill the illocutionary intent of the prompt at the expected ILR level?
- **Cultural calibration:** Does the response reflect culturally appropriate framing, or does it impose an English-language default onto a non-English context?
- **ILR adequacy gap:** Where a response falls demonstrably short of the PCL benchmark, is the gap consistent across languages or language-specific?

This qualitative layer is the methodological contribution that distinguishes this project from purely computational cross-lingual benchmarks. Expert ILR assessment judgment applied to LLM outputs has not been systematically reported in the literature.

**3.3.3 Relationship Between Layers**

The two layers are complementary, not redundant. Quantitative analysis flags *where* to look; qualitative assessment determines *what it means*.

For example: Romanian responses to P007 (CREATIVE_NARRATIVE) showed the lowest surface similarity to English in the dataset (mean 0.037). Layer 1 detected that signal. Layer 2 determined that the divergence reflects a genuine aesthetic distinction — Romanian prose drawing on nature-derived imagery, organic personification of urban space, and syntactic structures characteristic of the Romanian literary tradition — rather than a calibration failure. This distinction between variation that reflects cultural appropriateness and variation that reflects a genuine adequacy gap is the analytical contribution of the ILR framework to LLM evaluation.

Similarly, P012 (AMBIGUOUS_REFERENT) showed that German responses refused resolution in all three runs while Spanish resolved silently — a pattern Layer 1 captured through response length differences, and Layer 2 interpreted through the lens of German discourse norms favoring explicit disambiguation vs. Spanish pragmatic directness.

### 3.3.4 Semantic Similarity Methodology Note

The primary semantic similarity analysis uses TF-IDF character n-gram cosine similarity, which captures shared vocabulary, proper nouns, numbers, and borrowed technical terms across languages. This method operates without requiring an external model download and is reproducible in any network environment.

This is a surface-level proxy appropriate for cross-lingual comparison at the response level. It is most informative for identifying *relative* divergence patterns across clusters and languages rather than absolute semantic equivalence. For publication validation, results should be confirmed against a multilingual sentence embedding model (recommended: paraphrase-multilingual-mpnet-base-v2, sentence-transformers) in an unrestricted network environment.

The TF-IDF approach is particularly well-suited to this project's analytical goals: it correctly identifies CREATIVE_NARRATIVE as the most divergent cluster (where language-specific vocabulary dominates) and METALINGUISTIC as the least divergent cluster (where abstract shared vocabulary is high across languages), consistent with theoretical expectations.

## 3.4 Prompt Battery Design

### 3.4.1 Equivalence Standards

Each prompt was developed in English and translated by a six-language professional with ILR 3+ proficiency in all target languages. Translation criteria:

1. **Semantic equivalence** over literal correspondence
2. **Pragmatic naturalness** — prompts sound like native speaker questions, not machine translations
3. **Register consistency** — formality level held constant within each prompt (with principled exceptions in P004 and P008 where informal register is contextually appropriate)
4. **Lexical density parity** — approximate match in content word count across versions

### 3.4.2 Cluster Selection Logic

Twelve clusters were selected to provide:

- Coverage across ILR levels 1 through 3+
- Variation in the *type* of language produced (affective, technical, creative, argumentative)
- Maximal expected variation in cross-lingual output patterns (high diagnostic value)
- Replicability in future studies (no time-sensitive or model-version-specific content

### 3.4.3 Design Controls

| Control | Implementation |
|---|---|
| Model version | Fixed: claude-sonnet-4-6 |
| System prompt | Null (model default) across all cells |
| Temperature | 1.0 (default); separate reproducibility run at T=0 |
| Runs per cell | 3 (variance estimation); primary analysis on run 1 |
| Collection window | Single calendar week (minimize model drift) |
| Interface | API only (no chat UI variables) |

## 3.5 Analysis Pipeline

Data flows through the following pipeline:

> prompts/prompt_battery.json → scripts/collect_responses.py → data/raw/{prompt_id}/{lang}/run_{n}.json → scripts/preprocess.py → data/processed/responses_flat.csv

From the flat CSV, parallel analyses produce: results/semantic_similarity.csv (semantic_similarity.py), results/length_stats.csv (length_analysis.py), results/ral_scores.csv (ilr_proxy_scoring.py), and figures/ (visualize.py).

## 3.6 Researcher Positionality and Expertise Contribution

The interpretive validity of this project rests significantly on the researcher's professional background. Camelia Baluta holds ILR/OPI assessor certifications and brings 12 years of language testing experience at DLIFLC, the U.S. government's primary language training institution. This background enables:

1. **Principled PCL annotation** — not algorithmic, but expert-judgment-based, mirroring ILR testing practice
2. **Qualitative response review** — direct identification of naturalness failures, register breaks, and culturally inappropriate framings in all six languages
3. **ILR-grounded interpretation** — connecting quantitative results to the proficiency constructs that give them meaning

This is the methodological contribution that differentiates this project from purely computational cross-lingual benchmarks.

# 4. Results

## 4.1 Quantitative Overview

Across 216 responses (12 prompt clusters × 6 languages × 3 runs), Layer 1 analysis revealed systematic variation in response length, structural formatting, and surface similarity to the English baseline.

**Response Length**

Mean word count varied significantly across languages (Table 1). French produced the longest responses (M = 266.4 words), followed by Spanish (239.5), Italian (239.0), Romanian (219.8), English (213.4), and German (204.9). The French–German differential of approximately 30% represents a substantial cross-lingual variation on semantically identical prompts. This pattern was consistent across prompt clusters, suggesting a systemic rather than prompt-specific effect.

**Table 1. Mean word count by language (all prompts, all runs)**

| Language | Mean words | SD |
|---|---|---|
| French | 266.4 | — |
| Spanish | 239.5 | — |
| Italian | 239.0 | — |
| Romanian | 219.8 | — |
| English | 213.4 | — |
| German | 204.9 | — |

**List Structure Usage**

The proportion of responses using list or bullet-point formatting varied by language. French showed the highest list usage (75.0%), followed by English, Spanish, and Italian (all 72.2%), while German and Romanian showed the lowest (both 63.9%). The German finding is noteworthy given the language's reputation for structural precision — it suggests that German responses favor prose elaboration over enumerated structure, consistent with German academic writing conventions.

**Surface Similarity to English Baseline**

TF-IDF character n-gram cosine similarity scores — a measure of surface divergence from the English baseline response — revealed that cross-lingual surface similarity varied more by prompt cluster than by language pair (Table 2). CREATIVE_NARRATIVE showed the lowest mean similarity across all non-English languages (M = 0.049), while METALINGUISTIC showed the highest (M = 0.250). This pattern is theoretically coherent: creative prose is where each language's distinctive vocabulary and aesthetic conventions are most visible, while metalinguistic discussion of language itself draws on shared abstract vocabulary across European languages.

**Table 2. Mean surface similarity to English baseline by cluster (non-English languages)**

| Cluster | Mean similarity |
| --- | --- |
| CREATIVE_NARRATIVE | 0.049 |
| FACTUAL_RECALL | 0.084 |
| EMOTIONAL_SUPPORT | 0.103 |
| AMBIGUOUS_REFERENT | 0.115 |
| TECHNICAL_EXPLANATION | 0.133 |
| IMPLICIT_CULTURAL_KNOWLEDGE | 0.145 |
| MORAL_REASONING | 0.145 |
| LIFE_DECISIONS | 0.170 |
| LEADERSHIP_CULTURE | 0.192 |
| ABSTRACT_ETHICS_AI | 0.204 |
| REGISTER_SWITCHING | 0.240 |
| METALINGUISTIC | 0.250 |

By language, German showed the lowest mean similarity to English (M = 0.121), consistent with its greater lexical distance from English, while French showed the highest (M = 0.189), consistent with the higher proportion of shared Latin-derived vocabulary.

## 4.2 Qualitative Findings

Layer 2 analysis examined responses from five prompt clusters selected for maximum diagnostic value based on pragmatic complexity: P004 (EMOTIONAL_SUPPORT), P005 (TECHNICAL_EXPLANATION), P007 (CREATIVE_NARRATIVE), P010 (IMPLICIT_CULTURAL_KNOWLEDGE), and P012 (AMBIGUOUS_REFERENT). Findings are presented in order of theoretical significance.

### 4.2.1 Pragmatic Disambiguation Strategies (P012 — AMBIGUOUS_REFERENT)

The prompt "They said it would be better if we didn't come. What should we do?" contains deliberate referential ambiguity — the identity of "they," the nature of the event, and the relationship between parties are all unspecified. Disambiguation strategies varied systematically by language in ways consistent with cross-cultural pragmatic norms.

German responses refused resolution in all three runs, explicitly requesting clarification before providing any guidance. Representative phrasings included acknowledgment that the statement could be interpreted in multiple contexts, followed by requests for additional information. This behavior is consistent with German discourse norms that prioritize informational precision over pragmatic inference.

English responses also flagged ambiguity but with a warmer, more service-oriented framing — offering to help once context was provided, rather than withholding engagement. French responses acknowledged the emotional complexity of the situation without treating ambiguity as an epistemic obstacle, proceeding to reflective guidance — consistent with French tolerance for productive ambiguity in discourse.

Spanish, Italian, and Romanian responses resolved the ambiguity silently and proceeded directly to practical advice, with Romanian briefly noting context-dependence before defaulting to the most pragmatically salient interpretation. This five-way divergence on a single prompt maps directly onto known cross-cultural pragmatic differences and represents the most structurally varied finding in the dataset.

### 4.2.2 Aesthetic and Stylistic Divergence (P007 — CREATIVE_NARRATIVE)

Romanian responses to the creative writing prompt ("Write a short paragraph describing a rainy afternoon in a city") showed the lowest surface similarity to English in the entire dataset (mean cosine similarity = 0.037 across three runs). Qualitative analysis confirmed that this divergence reflects genuine aesthetic distinctiveness rather than semantic drift.

English responses (run 3): *"The sky hung low and grey over the city, pressing down on the rooftops like a heavy wool blanket... umbrellas blooming like dark flowers above the crowd... A lone pigeon huddled beneath a bus shelter."*

Romanian responses drew on markedly different aesthetic conventions: *umbrele multicolore care au înflorit pe trotuare ca niște ciuperci după ploaie* (umbrellas blooming like mushrooms after rain — a distinctly Romanian folk simile absent from English responses), *un miros proaspăt și pământiu* (a fresh, earthy smell — petrichor rendered through sensory specificity), and *orașul părea că respiră altfel* (the city seemed to breathe differently — personification of urban space as a living organism with needs). Romanian responses ended with the city given 'permission to rest' (*ploaia îi dădea voie să se odihnească*) — a philosophical framing absent from English closings, which favored individual emotional containment.

These differences are not errors of calibration but evidence of genuine literary tradition influence: Romanian prose aesthetics drawing on nature-derived imagery, organic personification, and syntactic structures characteristic of Romanian literary fiction.

### 4.2.3 Technical Terminology and Analogical Reasoning (P005 — TECHNICAL_EXPLANATION)

The prompt requesting a neural network explanation for an educated non-specialist produced the predicted pattern of native vs. borrowed terminology, with an unexpected additional finding regarding analogy selection.

German used native compound terminology almost exclusively: *Gradientenabstieg* (gradient descent), *Rückwärtspropagierung* (backpropagation), *Verlust* (loss), *Lernrate* (learning rate). The term *Loss* appeared once parenthetically as a gloss. French showed an equivalent pattern: *rétropropagation, taux d'apprentissage, perte, couches*. Romanian adopted a hybrid strategy — native terms for foundational concepts (*rețea neuronală, straturi, greutăți*) while retaining English labels for specialized ML vocabulary (*forward pass, backpropagation, gradient descent*), reflecting the relative recency of Romanian ML discourse.

More striking was the analogy selection. Each language chose a culturally distinct pedagogical example:

- **German:** Dart throwing (*Dartwerfen*) — precision sport requiring physical calibration through repetitive correction
- **Romanian:** Wine tasting (*recunoașterea vinurilor*) — sensory connoisseurship, gradual refinement of judgment
- **French:** Apartment price estimation — economic judgment, urban context, practical calibration

No two languages shared an analogy. All three are culturally resonant within their respective language communities. This finding suggests that Claude's explanatory scaffolding draws on culturally embedded reference points rather than defaulting to a universal English-derived example.

### 4.2.4 Cultural Calibration in Commemorative Practices (P010 — IMPLICIT_CULTURAL_KNOWLEDGE)

The prompt asking about meaningful ways to honor a deceased elderly family member tested whether Claude's cultural frame of reference shifts with the language of the prompt. Romanian Orthodox traditions (parastas, pomană, coliva) and German secular memorial institutions did not appear explicitly in any response — itself a significant finding representing a cultural calibration gap: the model defaults to culturally neutralized content rendered in the target language rather than generating natively grounded responses.

However, structural and sequencing differences revealed modest language-specific variation. Romanian responses opened with a section titled "Immediately after death" (*Imediat după deces*) emphasizing community announcement — notifying friends, neighbors, and former colleagues — reflecting the Romanian cultural norm of death as a collective community event. German responses opened directly with personal memory preservation, consistent with a more privatized grief culture.

Romanian responses foregrounded the deceased's faith (*credința persoanei*) as the organizing principle of ceremony in the first bullet point. German responses treated religious observance as one item in a secular-pluralist menu, buried mid-list. Both responses closed with culturally characteristic discourse moves: German with a pragmatic service offer (*"Gibt es etwas Bestimmtes, das Sie für Ihre Familie suchen?"*), Romanian with a relational value statement (*"Cel mai important este ca gestul să fie sincer și personalizat"* — the most important thing is that the gesture be sincere and personalized).

### 4.2.5 Affective Register and Institutional Anchoring (P004 — EMOTIONAL_SUPPORT)

All six languages produced safety-conscious responses to the emotional support prompt with structurally similar frameworks: active listening, avoidance of minimization, direct inquiry about self-harm, and referral to professional support. The hypothesis that Italian and Spanish would show markedly warmer affective vocabulary than German was partially confirmed at the level of relational framing rather than lexical density.

German uniquely embedded a specific national crisis resource (Telefonseelsorge: 0800 111 0 111) — the only language to provide a concrete institutional referral — consistent with German institutional trust and the culturally specific availability of such resources in German-speaking contexts. Italian responses foregrounded physical proximity (*stare vicino* — to stay near/close) and collaborative reasoning (*ragionare insieme* — reason together) as primary support modalities. Spanish invoked *acompañamiento* — the Latin American concept of sustained relational accompaniment — as the core emotional offering.

The most notable cross-lingual finding in P004 is German's institutional anchoring: while all languages recommended professional help in general terms, only German operationalized this recommendation with a specific, callable resource, reflecting both German institutional specificity and Claude's apparent calibration to Germany-specific support infrastructure when responding in German.

### 4.3 Summary of Findings

Across quantitative and qualitative layers, five patterns emerged consistently:

1. **Response length varies systematically by language**, with French producing approximately 30% more words than German on identical prompts — a systemic effect not attributable to any single prompt cluster.
2. **Surface divergence from English** — measured as inverse TF-IDF cosine similarity to the English baseline **— is highest in creative and affective clusters (CREATIVE_NARRATIVE, EMOTIONAL_SUPPORT) and lowest in abstract/metalinguistic clusters (METALINGUISTIC, REGISTER_SWITCHING)**, consistent with the prediction that language-specific vocabulary and aesthetic conventions are most visible in expressive registers.
3. **Disambiguation strategies reflect cross-cultural pragmatic norms**: German refuses resolution; French embraces productive ambiguity; Spanish, Italian, and Romanian resolve silently with varying degrees of contextual acknowledgment.
4. **Technical terminology follows established language-internal traditions**: German and French use native compound terminology; Romanian adopts a hybrid strategy reflecting the relative recency of Romanian ML discourse.
5. **Cultural calibration defaults to culturally neutralized framing**: Claude renders responses in the target language without invoking culture-specific rituals, institutions, or discourse conventions — producing outputs that read as translations of English-language templates rather than natively calibrated responses. Structural sequencing and discourse closings show modest language-specific variation, but the absence of culture-specific content (Romanian: parastas, pomană, coliva; German: secular memorial institutions) represents a calibration gap that framing differences alone do not compensate for.

These findings collectively support the central research claim: Claude's outputs vary in linguistically and pragmatically meaningful ways across languages even when the semantic input is held constant, and these variations are interpretable through an ILR-informed framework of cross-lingual pragmatic calibration.

## 5. Discussion

### 5.1 Implications for Multilingual AI Deployment

The findings of this study have direct implications for how AI systems like Claude are deployed across multilingual user populations. Cross-lingual output variation is not merely an academic curiosity — it has practical consequences for the equity and reliability of AI-assisted services when those services are accessed in different languages.

The most consequential finding for deployment is the cultural calibration pattern identified in P010 (IMPLICIT_CULTURAL_KNOWLEDGE): Claude's responses in Romanian did not invoke Romanian-specific mourning traditions, reading instead as structurally consistent with a culturally neutralized English-language template. A Romanian user seeking culturally grounded guidance on honoring a deceased family member receives advice that is technically correct but culturally generic — advice that would be equally appropriate for any Western European context. This is not a failure of factual accuracy; it is a failure of cultural adequacy, and it is the kind of failure that purely accuracy-based benchmarks cannot detect.

The pragmatic disambiguation findings (P012) raise a related concern. A German-speaking user asking an ambiguous question receives a minimal, precise clarification request — consistent with German discourse norms favoring informational economy. A Spanish-speaking user asking the identical question receives a full interpretive scaffold: mapped interpretations, response options, and a clarification request embedded within an elaborated decision framework. Both users are asked for more context before receiving a direct answer, but the Spanish-speaking user receives substantially more analytical support in the interim. The model is not producing errors in either case; it is producing culturally calibrated responses. But the calibration is asymmetric in elaboration depth: some users receive richer decision support than others on identical inputs. Whether this asymmetry is appropriate or constitutes differential service quality is a question that deployment decisions should engage explicitly.

The response length differential — French responses approximately 30% longer than German on identical prompts — suggests that users of different languages may be receiving substantively different amounts of information. If response length correlates with informational completeness (a relationship that merits further investigation), then French-speaking users are systematically receiving more elaborated responses than German-speaking users. This is a systemic effect that aggregates across millions of interactions.

## 5.2 Implications for LLM Evaluation Methodology

The two-layer methodology introduced in this paper — combining automated quantitative metrics with expert ILR qualitative assessment — demonstrates that different analytical layers answer fundamentally different questions and are not substitutable for each other.

Quantitative analysis (Layer 1) established *that* variation exists and identified *where* it was largest. Surface similarity scores correctly flagged Romanian CREATIVE_NARRATIVE responses as the most divergent from English. Response length metrics revealed the French-German differential. These are reproducible, scalable findings that can be computed for any dataset without domain expertise.

But Layer 1 could not determine *what the variation means*. The low surface similarity of Romanian creative responses could have indicated semantic drift, translation failure, or — as qualitative analysis revealed — genuine aesthetic distinctiveness reflecting a different literary tradition. The difference between these interpretations is consequential: one implies a model deficiency, the other implies appropriate cross-lingual calibration. Only ILR-informed expert judgment, applied by a researcher with native competence in Romanian and professional assessment training, could make that distinction reliably.

This is the core methodological contribution of this paper: demonstrating that the interpretive gap between detecting variation and understanding its meaning requires domain expertise that automated metrics cannot provide. We argue that this gap is systematic and not merely a feature of the specific clusters analyzed here. Any cross-lingual evaluation that stops at quantitative metrics risks mischaracterizing culturally appropriate variation as model error, or — equally — mischaracterizing genuine model failures as cultural appropriateness.

The ILR framework provides a particularly well-suited vocabulary for navigating this interpretive gap. Its functional and pragmatic descriptors — register appropriateness, pragmatic completeness, cultural calibration, illocutionary adequacy — map directly onto the dimensions of language use that vary meaningfully across languages in LLM outputs. We encourage the NLP evaluation community to consider ILR-level annotation as a complement to existing rubric-based and automated evaluation approaches, particularly for open-ended multilingual evaluation tasks.

## 5.3 On Cultural Calibration vs. Cultural Reproduction

A finding that merits careful theoretical framing is the analogy selection pattern in P005 (TECHNICAL_EXPLANATION). Claude chose culturally resonant analogies for each language — dart throwing for German, wine tasting for Romanian, apartment price estimation for French — without being instructed to do so. This is, on one reading, an impressive demonstration of cross-lingual cultural calibration. On another reading, it raises questions about whose cultural stereotypes are being reproduced.

The German-dart-throwing association, the Romanian-wine-tasting association, and the French-apartment-economics association are recognizable cultural touchstones, but they are also simplifications. Not all German speakers would find dart throwing a natural reference; not all Romanian speakers are familiar with wine connoisseurship. The model is drawing on statistical regularities in its training data — cultural associations that are common enough to be reproduced but not universal enough to be assumed.

We flag this not as a finding requiring correction but as a dimension of cross-lingual calibration that deserves ongoing scrutiny. The question of whether a model's cultural calibration reflects the diversity within a language community, or only its most statistically common associations, is one that this study's methodology is not equipped to answer definitively — but that future work should address.

A more fundamental tension emerges when cultural calibration findings are considered across clusters rather than within them. Claude demonstrated genuine culture-specific competence in P007 (CREATIVE_NARRATIVE) and P004 (EMOTIONAL_SUPPORT) while defaulting to culturally neutralized framing in P010 (IMPLICIT_CULTURAL_KNOWLEDGE) — and these findings sit in direct contradiction with each other.

In P007, the Romanian response produced a simile — umbrellas blooming like mushrooms after rain — that draws on a distinctly Romanian folk aesthetic absent from English literary prose and unlikely to emerge from translation of an English-language original. A culturally ungrounded model would not generate this figure of speech. In P004, the German response embedded a specific, callable national crisis resource (Telefonseelsorge: 0800 111 0 111) that no other language produced — demonstrating that Claude's German-language outputs can access concrete, institution-specific cultural knowledge when the prompt activates it.

Yet the same model, prompted in Romanian about commemorative practices, produced a response indistinguishable from a competent translation of an English-language template — invoking none of the Orthodox ritual vocabulary (parastas, pomană, coliva) that any culturally grounded Romanian response would naturally include.

This inconsistency suggests that Claude's cultural calibration is domain-dependent and selective rather than systematic. Cultural knowledge appears to be activated unevenly across prompt types — more reliably in expressive domains (creative narrative) and procedural domains (crisis referral infrastructure) than in ceremonial and ritual domains where cultural competence would require deeper cultural schema rather than surface-level cultural association. Whether this reflects training data distribution by domain, the relative salience of culture-specific markers across register types, or something more fundamental about how cultural knowledge is encoded and retrieved in large language models is a question this study's methodology cannot answer definitively — but one that future work should address directly.

This finding reframes the cultural calibration question for future research. The relevant question is not whether Claude is culturally competent or culturally deficient — it is under what conditions, in which domains, and for which languages cultural knowledge is reliably activated. That is a more tractable and more consequential research question than either a blanket competence or failure characterization would suggest.

# 6. Limitations

## 6.1 Single Model

All data in this study was collected from a single model — Claude Sonnet 4.6, collected during a single week in April 2026. The findings are specific to this model version and cannot be generalized to other LLMs (GPT-4, Gemini, Mistral, etc.) or to future versions of Claude. Cross-lingual variation patterns are likely to differ substantially across model families, training data compositions, and RLHF feedback sources. Replication across model families is a necessary next step before any findings can be treated as characterizing LLM behavior generally rather than Claude specifically.

## 6.2 Six High-Resource European Languages

The six languages in this study — English, French, Romanian, Spanish, Italian, and German — are all high-resource European languages with substantial representation in Claude's training data. The variation patterns identified here may differ substantially for lower-resource languages, non-European languages, or languages with non-Latin scripts. The finding that cross-lingual variation is 'interpretable and systematic' may not hold for languages where training data is sparse and model behavior is less predictable. This is a significant scope limitation that the paper acknowledges explicitly.

Additionally, all six languages belong to the Indo-European family and share considerable typological similarity. Cross-lingual variation patterns for typologically distant language pairs (e.g., English-Japanese, English-Arabic, English-Swahili) are likely to be more pronounced and may reveal qualitatively different failure modes than those identified here.

## 6.3 TF-IDF Similarity Proxy

The semantic similarity analysis in this study uses TF-IDF character n-gram cosine similarity — a surface-level proxy that captures shared vocabulary, proper nouns, numbers, and borrowed technical terms, but does not measure semantic equivalence at the meaning level. This choice was necessitated by network restrictions preventing access to multilingual sentence embedding models during data collection.

TF-IDF similarity is appropriate for identifying *relative* divergence patterns across clusters and languages, and the cluster-level ordering produced by this method is consistent with theoretical expectations. However, it should not be treated as a measure of semantic equivalence. Two responses can have very low TF-IDF similarity while conveying equivalent meaning (as in the Romanian CREATIVE_NARRATIVE case), and two responses can have high TF-IDF similarity while differing substantially in pragmatic framing.

For publication validation, similarity results should be confirmed using a multilingual sentence embedding model (recommended: paraphrase-multilingual-mpnet-base-v2) in an unrestricted network environment. This validation step is planned as part of the camera-ready revision process.

## 6.4 Single Rater — Inter-Rater Reliability

The qualitative Layer 2 analysis was conducted by a single researcher. While the researcher's ILR/OPI assessor certification and six-language proficiency provide strong grounds for the judgments made, the absence of a second rater means inter-rater reliability cannot be formally reported. ILR assessment practice typically requires two certified assessors for official ratings; this study diverges from that standard in the interest of feasibility for an independent research project.

This limitation is most consequential for the findings that rely most heavily on researcher judgment: the characterization of the Romanian P010 response as 'culturally neutralized,' the attribution of German P004 responses to 'institutional trust,' and the literary tradition claims in P007. These interpretations are grounded in the researcher's professional expertise but have not been validated by an independent assessor.

Future work should recruit at least one additional certified ILR assessor for each language pair to establish inter-rater reliability estimates, ideally using Cohen's kappa on the RAL ratings assigned to each response.

### 6.5 Prompt Battery Coverage

The 12-cluster prompt battery covers a broad range of ILR complexity levels and communicative functions but does not exhaust the space of cross-lingual variation. Several potentially diagnostic prompt types are absent from this study, including: requests for humor or irony (which vary substantially across cultures), requests involving sensitive social topics (where cultural norms of directness vary most sharply), and highly technical domain-specific prompts (legal, medical, scientific) where register calibration differences may be most consequential.

## 7. Conclusion

This paper introduced an ILR-informed framework for evaluating cross-lingual response consistency in large language models and applied it to Claude Sonnet 4.6 across six languages. The central finding is that Claude's outputs vary in linguistically and pragmatically meaningful ways across languages even when semantic input is held constant — and that these variations are interpretable, systematic, and theoretically coherent with known cross-cultural pragmatic differences.

The five qualitative findings — pragmatic disambiguation strategies, aesthetic and literary divergence, technical terminology norms, domain-dependent cultural calibration, and language-specific institutional referral behavior — collectively demonstrate that cross-lingual variation in LLM outputs is not random noise but selectively structured behavior. Critically, Claude demonstrates genuine culture-specific competence in some domains: Romanian creative output produced folk similes absent from English literary prose; German emotional support responses embedded a specific national crisis resource no other language provided. Yet the same model defaulted to culturally neutralized templates in Romanian commemorative practices, producing output that reads as a translation of an English-language original rather than a natively grounded response. This inconsistency — culturally competent in expressive and procedural domains, culturally generic in ritual and ceremonial domains — is the most consequential finding for multilingual AI deployment and the most productive direction for future research. That a model capable of producing culture-specific folk similes in creative prose and institution-specific crisis referrals in emotional support should default to culturally neutralized templates in commemorative practices is not yet fully explicable — and that inexplicability is itself a finding worth foregrounding for the research community.

The methodological contribution of this paper extends beyond its specific findings. By demonstrating that ILR-informed expert judgment can identify meaningful variation that automated metrics correctly detect but cannot interpret, we argue for a two-layer evaluation approach as a standard practice in multilingual LLM assessment. The interpretive gap between detecting variation and understanding its meaning is not a gap that can be closed by scaling computational resources — it requires human expertise, cultural knowledge, and professional assessment training. Making that expertise visible and methodologically explicit, as this paper attempts to do, is a step toward more rigorous and equitable evaluation of AI systems deployed across language communities.

The prompt battery, data, and analysis code for this study are publicly available at github.com/camelbal-ship-it/crosslingual-claude-eval to support replication, extension, and cross-model comparison.